\documentclass[opre,nonblindrev]{informs3_homepage}

\OneAndAHalfSpacedXI

\usepackage{subcaption}
\usepackage{xcolor}
\definecolor{pastelBlue}{rgb}{0.0,0.4,0.7}
\usepackage{hyperref}
\hypersetup{
colorlinks=true,
allcolors = pastelBlue  
}

\usepackage{natbib}
 \bibpunct[, ]{(}{)}{,}{a}{}{,}%
 \def\bibfont{\footnotesize}%
\TheoremsNumberedThrough     
\ECRepeatTheorems

\EquationsNumberedThrough    

\usepackage{float}
\usepackage{amsmath}
\usepackage{algorithm2e}
\RestyleAlgo{ruled}
\LinesNumbered
\let\oldnl\nl
\newcommand{\nonl}{\renewcommand{\nl}{\let\nl\oldnl}}

\usepackage{algpseudocode}
\usepackage{ifthen}
\usepackage{enumitem}
\usepackage{cases}
\usepackage{mathtools}
\usepackage{amssymb}
\usepackage{graphicx}
\usepackage{empheq}
\usepackage{tikz}
\usetikzlibrary{shapes.geometric}
\usetikzlibrary{shapes}
\usetikzlibrary{arrows}
\usetikzlibrary{calc,positioning}
\usepackage{bm}
\usepackage{dsfont}
\usepackage{booktabs}

\usepackage{macros}

\begin{document}

\RUNTITLE{On Evolution-Based Models for Experimentation Under Interference}

\TITLE{\large On Evolution-Based Models for Experimentation Under Interference}

\ARTICLEAUTHORS{%
\AUTHOR{Sadegh Shirani~~~~~~~and~~~~~~~Mohsen Bayati}
\AFF{Graduate School of Business, Stanford University}
}
\ABSTRACT{%
Causal effect estimation in networked systems is central to data-driven decision making. In such settings, interventions on one unit can spill over to others, and in complex physical or social systems, the interaction pathways driving these interference structures remain largely unobserved. We argue that for identifying population-level causal effects, it is not necessary to recover the exact network structure; instead, it suffices to characterize how those interactions contribute to the evolution of outcomes. Building on this principle, we study an evolution-based approach that investigates how outcomes change across observation rounds in response to interventions, hence compensating for missing network information. Using an exposure-mapping perspective, we give an axiomatic characterization of when the empirical distribution of outcomes follows a low-dimensional recursive equation, and identify minimal structural conditions under which such evolution mappings exist. We frame this as a distributional counterpart to difference-in-differences. Rather than assuming parallel paths for individual units, it exploits parallel evolution patterns across treatment scenarios to estimate counterfactual trajectories. A key insight is that treatment randomization plays a role beyond eliminating latent confounding; it induces an \emph{implicit sampling from hidden interference channels}, enabling consistent learning about heterogeneous spillover effects. We highlight causal message passing as an instantiation of this method in dense networks while extending to more general interference structures, including influencer networks where a small set of units drives most spillovers. Finally, we discuss the limits of this approach, showing that strong temporal trends or endogenous interference can undermine identification. 
}


\maketitle

\vspace{-1.25cm}

\section{Introduction}
\label{sec:Intro}
Analyzing network data is central to scientific fields, with broad applications spanning operations research, economics, and social sciences \citep{jackson2008social,johari2022experimental}. These analyses are inherently challenging due to complex dependencies shaped by latent and observed network structures \citep{kolaczyk2014statistical}. Estimating causal effects in networked systems presents a particularly challenging case \citep{hudgens2008toward,ugander2013graph,eckles2016design}. This complexity arises because, in addition to intricate network dependencies, researchers must face the fact that we can observe outcomes under only a single realized scenario \citep{holland1986statistics}.

Interference induced by underlying network structures violates the stable unit treatment value assumption (SUTVA), necessitating methods that can disentangle causal relationships in interconnected settings \citep{imbens2015causal}. Within the growing literature on experimental design under interference, one line of research investigates the use of outcome observations collected over time. Specifically, \cite{li2022network} highlight the utility of additional longitudinal data for inference and the complications introduced by temporal interference \citep{glynn2020adaptive}.

Recent contributions have exploited temporal structures and observations to enhance estimation under interference \citep{farias2022markovian,hu2022switchback,bojinov2023design,li2023experimenting,ni2023design,boyarsky2023modeling,han2024population,mukaigawara2025spatiotemporal,jia2025clustered}. 
Collectively, these studies demonstrate that temporal dynamics can be utilized effectively to mitigate bias and improve statistical inference in the presence of interference, but they typically retain a unit‑level or network‑level description of interference.

This temporal approach has also been used in broader network problems to detect connections and structural properties. Systematic perturbations followed by observing the resulting outcome trajectories can provide information about network topology \citep{timme2014revealing}. For example, observing transient responses to targeted or random perturbations enables the identification of causal relations among network nodes \citep{casadiego2017model,nitzan2017revealing,stepaniants2020inferring}.

Recently, \cite{shirani2024causal} introduced causal message passing (CMP) for estimating treatment effects in networked populations. Building on the approximate message passing methodology \citep{donoho2009message,bayati2011dynamics}, CMP introduces the concept of \textbf{experimental state evolution (ESE)}. ESE captures how outcome distributions evolve over time, with this temporal evolution characterized through mathematical functions called \textbf{ESE mappings}.

The derivation of ESE mappings and their application to treatment effect estimation have been studied under certain structural assumptions on outcome dynamics and interference patterns \citep{shirani2024causal,bayati2024higher,shirani2025can}. These papers derive ESE as a consequence of particular unit‑level outcome models and random network assumptions. However, it remains unclear when such evolution‑based descriptions exist, how they relate to classical exposure mappings for interference \citep{manski2013identification,aronow2017estimating}, and under what structural conditions they support identification of population‑level counterfactual trajectories without reconstructing the underlying network. In this paper, we reverse the perspective. Instead of starting from a detailed outcome and network model and characterizing limiting distributions, we take ESE itself as the primitive object and ask:
\begin{enumerate}
    \item Under what assumptions on potential outcomes and exposure mappings does an ESE representation exist at the population level?

    \item When such a representation exists, when does it suffice for identification of counterfactual evolutions, and how does it relate to familiar assumptions such as parallel trends?

    \item How can partial structural knowledge, such as the presence of a small set of influencers or known clusters, be incorporated without requiring full network reconstruction?

    \item In which environments do evolution-based approaches fail, regardless of the estimator?
\end{enumerate}

Our contributions are as follows. First, we provide an axiomatic treatment of the ESE framework. Using a potential‑outcome evolution model and time‑varying exposure mappings, we derive minimal regularity conditions under which population outcome distributions follow an ESE recursion driven by treatment assignments. Second, we show how ESE underpins an evolution‑based estimation strategy and clarify its connection to difference‑in‑differences: under a distributional parallel evolution condition, counterfactual trajectories can be constructed by repeatedly propagating from common pre‑treatment baselines, providing a distributional analogue of the parallel‑trends assumption. This generalizes the evolution‑based viewpoint beyond the dense network regimes considered by CMP and moves beyond the limitations of these methods. As concrete illustrations, we demonstrate how the approach extends to core-periphery settings where a small but known set of ``influencers'' exert disproportionate spillover effects. Finally, we show intrinsic limitations of evolution‑based methods: we show that strong time trends or treatment‑dependent exposure mechanisms violate the stability conditions required for ESE, and document empirically how such violations degrade evolution‑based estimators.

\begin{figure}
    \centering
    \includegraphics[width=0.9\linewidth]{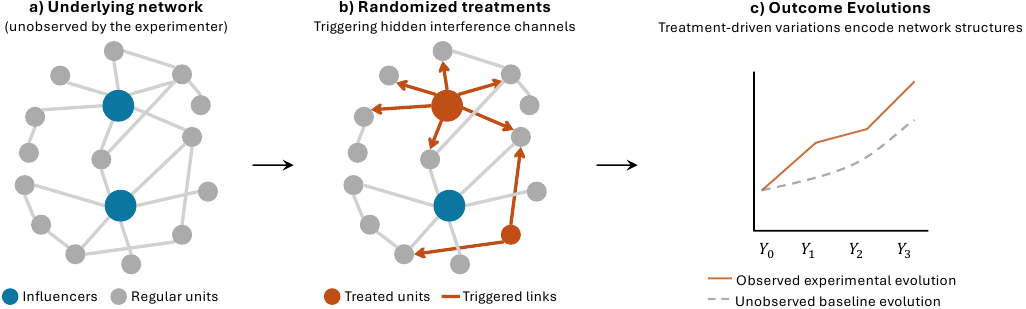}
    \caption{Randomized treatments implicitly sample hidden interference channels in a network with two influencers.}
    \label{fig:FigOne}
\end{figure}

A key insight is that treatment randomization addresses two separate challenges. The first concerns \emph{unobserved characteristics}: random assignment makes treated and control units comparable on average, thereby removing confounding. The second, newly emphasized here, concerns \emph{unobserved network structure}. Randomization injects structured variation into the hidden interference channels by triggering a representative subset of the links through which spillovers operate. By tracking how outcomes evolve across these perturbations, we effectively sample the latent pathways that transmit treatment effects. This \emph{implicit sampling} mechanism enables identification of population-level causal effects without reconstructing the underlying network (Figure~\ref{fig:FigOne}).

The remainder of the paper proceeds as follows. Section~\ref{sec:problem_formulation} introduces experimental setup along with an illustrative example that motivates the foundation of our approach. Section~\ref{sec:SE} establishes a set of minimal assumptions for the existence of ESE mappings. Section~\ref{sec:interference} analyzes ESE mappings through the lens of exposure mappings. Section~\ref{sec:DPT} explains when outcome evolutions can be used for estimation. Finally, Section~\ref{sec:conclusion} discusses limitations and delineates the scope of the framework.

\section{Problem Statement}
\label{sec:problem_formulation}

We consider a set of $\UN$ units indexed by $i=1,\ldots,\UN$, and suppose that all units can be observed across $\TH$ rounds indexed by $t = 1, \ldots, \TH$. For simplicity, we focus on binary treatment interventions and denote by $\treatment{i}{t}$ the treatment applied to unit~$i$ in round $t$. We assume that $\treatment{i}{t} \sim \text{Bernoulli}(\expr_t)$ for some $0 \leq \expr_t \leq 1$, where $\treatment{i}{t}=1$ indicates that unit~$i$ receives the treatment in round $t$ and $\treatment{i}{t}=0$ otherwise. We denote the treatment assignments across all units in round $t$ by $\Vtreatment{}{t} := (\treatment{1}{t}, \ldots, \treatment{\UN}{t})^\top$, and define $\Mtreatment{t} := [\Vtreatment{}{1}| \ldots | \Vtreatment{}{t}]$ as the collective treatment assignment up to round~$t$. We then use $\observedtreatment{i}{t}$, $\Vobservedtreatment{}{t}$, and $\Mobservedtreatment{t}$ to denote generic realizations of $\treatment{i}{t}$, $\Vtreatment{}{t}$, and $\Mtreatment{t}$, respectively.

Letting $\outcome{i}{0}$ denote the initial outcome of unit~$i$ prior to any treatment administration, we follow the Neyman-Rubin causal framework \citep{rubin1978bayesian,imbens2015causal}. Precisely, we assume that for any feasible treatment assignment $\Mobservedtreatment{t} \in \{0,1\}^{\UN \times t}$, the potential outcome of unit~$i$ in round~$t$ exists and denote it by $\outcome{i}{t}(\Mobservedtreatment{t}) \in \R$. Then, we denote the vector of potential outcomes in round~$t$ by
$\Voutcome{}{t}(\Mobservedtreatment{t}) := \big(\outcome{1}{t}(\Mobservedtreatment{t}), \ldots, \outcome{\UN}{t}(\Mobservedtreatment{t})\big)^\top$
and the panel of outcomes by
$\Moutcome{}(\Mobservedtreatment{\TH}) := \big[\Voutcome{}{0} \big| \Voutcome{}{1}(\Mobservedtreatment{1}) \big| \ldots \big| \Voutcome{}{\TH}(\Mobservedtreatment{\TH})\big]$.
The experimentation procedure consists of assigning treatments according to $\Mtreatment{\TH}$ and recording the outcome panel $\Moutcome{}(\Mtreatment{\TH})$. Then, motivated by population-level causal estimands,%
\footnote{A well-studied example of population-level estimands is the total (or global) treatment effect, which contrasts the sample mean of outcomes under universal treatment with those under no treatment \citep{yu2022estimating}.}
we focus on aggregated outcomes: \emph{how changes in $\Mtreatment{\TH}$ impact the distribution of observed outcomes $\Moutcome{}(\Mtreatment{\TH})$.} 

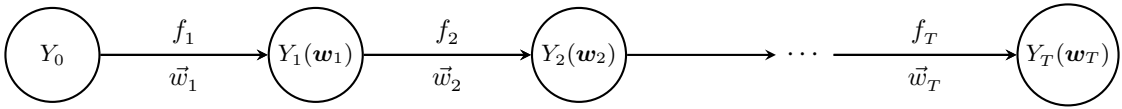
\begin{figure}[htbp]
\centering
\begin{tikzpicture}[
    scale=0.05,
    node distance=3.5cm,
    auto,
    thick,
    >=stealth,
    outcome/.style={circle, draw, minimum size=1.25cm, inner sep=2pt},
    arrow/.style={->, thick}
]
\node[outcome] (Y0) {\footnotesize$\outcome{}{0}$};
\node[outcome, right of=Y0] (Y1) {\footnotesize$\outcome{}{1}(\Mobservedtreatment{1})$};
\node[outcome, right of=Y1] (Y2) {\footnotesize$\outcome{}{2}(\Mobservedtreatment{2})$};
\node[right of=Y2, xshift=-0.5cm] (dots) {$\cdots$};
\node[outcome, right of=dots] (YT) {\footnotesize$\outcome{}{\TH}(\Mobservedtreatment{\TH})$};
\draw[arrow] (Y0) -- (Y1) node[midway, above, font=\small] {$\SEfunction{}{1}$};
\draw[arrow] (Y0) -- (Y1) node[midway, below, font=\small] {$\Vobservedtreatment{}{1}$};
\draw[arrow] (Y1) -- (Y2) node[midway, above, font=\small] {$\SEfunction{}{2}$};
\draw[arrow] (Y1) -- (Y2) node[midway, below, font=\small] {$\Vobservedtreatment{}{2}$};
\draw[arrow] (Y2) -- (dots) node[midway, above, font=\small] {};
\draw[arrow] (dots) -- (YT) node[midway, above, font=\small] {$\SEfunction{}{\TH}$};
\draw[arrow] (dots) -- (YT) node[midway, below, font=\small] {$\Vobservedtreatment{}{\TH}$};
\end{tikzpicture}
\caption{Experimental State Evolution: sequential transformation of outcome distributions via ESE mappings.}
\label{fig:ESE_evolution}
\end{figure}

We now provide a formal definition of the sequence of ESE mappings $\SEfunction{}{1}, \ldots, \SEfunction{}{\TH}$. For each round~$t$, we define random variables $\outcome{}{t-1}(\Mobservedtreatment{t-1}), \; \outcome{}{t}(\Mobservedtreatment{t}) \in \R$ that follow the same distributions as the empirical distribution of the elements of the outcome vectors $\Voutcome{}{t-1}(\Mobservedtreatment{t-1}), \; \Voutcome{}{t}(\Mobservedtreatment{t}) \in \R^{\UN}$, respectively. The ESE mapping $\SEfunction{}{t}$ then establishes a functional relationship from $\outcome{}{t-1}(\Mobservedtreatment{t-1})$ to $\outcome{}{t}(\Mobservedtreatment{t})$.
This sequential evolution process is illustrated in Figure~\ref{fig:ESE_evolution}. Indeed, $\SEfunction{}{t}$ captures the mechanism by which past outcomes $\outcome{}{t-1}(\Mobservedtreatment{t-1})$ evolve in response to current treatments $\Vobservedtreatment{}{t}$ to generate new outcomes $\outcome{}{t}(\Mobservedtreatment{t})$.
The following example illustrates how the network structure contributes to this evolution.

\vspace{-1cm}
\subsection*{An Illustrative Example: Outcomes Evolution Encode Network Structure!}
\begin{figure}
    \centering
    \includegraphics[width=0.9\linewidth]{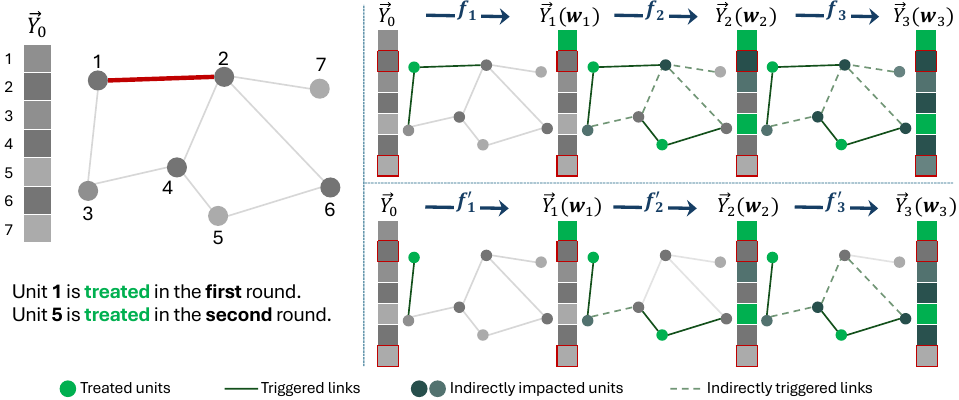}
    \caption{Outcomes evolution: treatments trigger interference channels both directly and indirectly, shaping the next column of outcomes. The columns show outcomes with (top) and without (bottom) a direct link between units~1 and~2. Even without observing the network, the outcome evolutions (for units~2 and~7) reveal differences in interference structures, implying distinct ESE mappings across the two cases.}
    \label{fig:evolution}
\end{figure}

Consider a community of seven individuals, some of whom interact regularly, as shown on the left side of Figure~\ref{fig:evolution}. We aim to share new information about the benefits of regular exercise and study its impact on individuals' behavior by measuring each person’s daily workout time. In this context, the initial outcome vector~$\Voutcome{}{0}$ denotes the activity levels of individuals before the experiment begins.

To conduct the experiment, we share the information with a randomly selected group of individuals. These interventions not only impact the workout time of those treated units but can also alter the content of their regular conversations with their friends. For example, two friends may already speak daily about unrelated topics. Once one of them receives the new information, their conversation may shift to include the benefits of exercise. This shift marks their link as \textbf{triggered}: the connection does not change structurally, but its functionality may change relative to the baseline, allowing the treatment-related information to flow through it.

In subsequent days, some of those newly informed friends may pass the information to their own contacts, generating additional rounds of triggered interactions. Although we do not observe these triggered links or the conversations that carry them, we do observe their aggregate effect in how the outcome vector changes from one day to the next. By comparing each column of outcomes with the previous one, we see which parts of the population respond more strongly, revealing the footprint of the underlying interference pathways. This is the key insight: the evolution of outcomes encodes the network effect; in fact, \emph{the treatment-driven variations in outcome columns provide \textbf{indirect} but observable evidence of the unobserved interference structure.}

Figure~\ref{fig:evolution} illustrates this idea by comparing two otherwise identical seven-person networks: one in which Individuals~1 and~2 are friends (top panel) and one in which they are not (bottom panel). In both cases, Individuals~1 and~5 are treated in the first and second rounds, respectively, and outcomes are shown as columns, with color intensity representing workout activity. When 1 and~2 are friends, the influence of Individual~1 spreads through their connection, leading to stronger responses in subsequent rounds. When they are not connected, this channel is absent, and the spillover pattern weakens (see the outcomes of Individuals~2 and~7). The differences in these evolving outcome patterns reveal the presence or absence of hidden connections.

In real-world experiments, networks are large and heterogeneous, and their structure is rarely observed. Randomization makes it possible to handle this complexity through an \textbf{implicit sampling mechanism}. When we randomly choose whom to inform about the benefits of exercise, we also randomly \emph{trigger} the social links connected to those individuals by shifting the content of their interactions. Each treated person activates a distinct local neighborhood, and taken together these activations form a representative sample of the hidden connections in the community. The resulting randomized triggers shape how outcomes evolve from one round to the next, allowing us to infer properties of unobserved interference from observed dynamics. Because it is infeasible to track all such variations one by one, we instead characterize their aggregate through the evolution of outcome distribution. ESE mappings formalize this idea by modeling how outcome distributions change as treatments trigger interaction links, translating temporal patterns in the data into information about the underlying interference structure.

\section{Experimental State Evolution}
\label{sec:SE}
We begin by introducing two new notations. In addition to treatments and outcomes, each unit~$i$ also has a covariate vector $\Vcovar{i}{t} \in \R^{\covardim}$ in each round~$t$, for some positive integer $\covardim$. This vector can incorporate unit-specific characteristics, such as age and gender for individuals, as well as time-dependent indicators, including the round number $t$. These covariates serve to characterize the heterogeneity across units and over time throughout the experimentation period.

Furthermore, to account for potential interference between experimental units, we rely on the concept of exposure mappings \citep{manski2013identification,aronow2017estimating}. Specifically, we let $\Vexposure{i}{t}(\Mobservedtreatment{t})$ be the \emph{exposure} vector taking values in $\exposurespace$, for some positive integer $\exposuredim$. That means, for a treatment allocation $\Mobservedtreatment{t}$, the elements of $\Vexposure{i}{t}(\Mobservedtreatment{t})$ encompass the effective influence received by unit $i$ in round~$t$ from all other experimental units. The following assumption formalizes this.
\begin{assumption}[Exposure vectors]
    \label{asmp:exposure_vecs}
    For any treatment assignment $\Mobservedtreatment{t}$, there exist exposure vectors $\Vexposure{1}{t}(\Mobservedtreatment{t}), \ldots, \Vexposure{\UN}{t}(\Mobservedtreatment{t})$ that summarize the interference effects received by each unit in round~$t$.
\end{assumption}

We proceed by characterizing the unit-level outcome evolution. For simplicity, we assume that the outcome of unit $i$ in round $t$ (denoted by $\outcome{i}{t}(\Mobservedtreatment{t})$) depends only on its most recent outcome $\outcome{i}{t-1}(\Mobservedtreatment{t-1})$, the current treatment $\observedtreatment{i}{t}$, as well as the the covariates $\Vcovar{i}{t}$ and exposure vectors $\Vexposure{i}{t}(\Mobservedtreatment{t})$. 

\noindent
This assumption defines a minimal memory model that facilitates tractable analysis. Our framework extends to more complex dynamics that involve more memory terms or full treatment trajectories, without requiring structural changes to the analytical setup. We formalize our potential outcome specification through the following assumption.
\begin{assumption}[Potential outcome evolution]
    \label{asmp:unit-level_dynamics}
    We assume that all unit outcomes evolve following a similar mathematical rule and define the generic unit-level evolution through a function~$\outcomefunction{}$ such that given initial pretreatment outcomes $\Voutcome{}{0}$, we have
    \begin{align}
        \label{eq:outcome_specification}
        \outcomefunction{}: 
        \treatmentspace \times \outcomespace \times \covarspace \times
        \exposurespace
        \mapsto \outcomespace,
        \quad
        \outcome{i}{t}(\Mobservedtreatment{t})
        :=
        \outcomefunction{}
        \big(
        \observedtreatment{i}{t},
        \outcome{i}{t-1}(\Mobservedtreatment{t-1}),
        \Vcovar{i}{t};
        \Vexposure{i}{t}(\Mobservedtreatment{t})
        \big).
    \end{align}
\end{assumption}
Equation~\eqref{eq:outcome_specification} presents the unit-level outcome dynamics. Importantly, evolution functions are assumed to be homogeneous across units and any heterogeneity arises through covariates $\covar{i}{t}$, and exposure vectors $\Vexposure{i}{t}(\Mobservedtreatment{t})$. Consequently, two units with identical covariates and exposures will follow the same evolution path if initialized identically and share the exact same treatment trajectory.

Note that Eq.~\eqref{eq:outcome_specification} characterizes the outcome evolution process, capturing how outcomes in each round depend on the previous round by incorporating $\outcome{i}{t-1}(\Mobservedtreatment{t-1})$ on the right-hand side. This differs from the outcome generating process that maps treatments to potential outcomes without incorporating $\outcome{i}{t-1}(\Mobservedtreatment{t-1})$ \citep{abadie2025causal}. Our goal is to demonstrate how aggregate outcomes evolve according to these unit-level dynamics. To this end, we consider a probability space $(\Omega, \F, \P)$, with $\Omega$ representing the sample space, $\F$ the sigma-algebra of events, and $\P$ the probability measure.
\begin{assumption}[Convergent potential outcomes]
    \label{asmp:empirical_distributions}
    For any treatment allocation $\Mobservedtreatment{\TH}$, the potential outcome panel $\Moutcome{}(\Mobservedtreatment{\TH})$ exists and satisfies the following condition.%
    \footnote{For $(x^1,\dots,x^\ell)$, the empirical distribution is $\frac{1}{\ell} \sum_{j=1}^\ell \delta_{x^j}$, where $\delta_{x}$ is the Dirac measure at $x$. Weak convergence of probability measures $\mu_n$ to $\mu$ means $\int \psi\, d\mu_n \to \int \psi\, d\mu$ for all bounded continuous functions $\psi$.}
    For each round~$t$, let $\empdist{t;\Mobservedtreatment{t}}{N}$ denote the empirical distributions of
    $\big\{ \big( \observedtreatment{i}{t}, \outcome{i}{t-1}(\Mobservedtreatment{t-1}), \Vcovar{i}{t}, \Vexposure{i}{t}(\Mobservedtreatment{t}), \outcome{i}{t}(\Mobservedtreatment{t}) \big) \big\}_{i = 1}^\UN$
    that converges weakly, as $\UN \to \infty$, to a probability distribution $\limdist{t;\Mobservedtreatment{t}}{}$.
\end{assumption}
This assumption establishes the regularity conditions necessary for asymptotic analysis in our experimental setting with evolving outcomes. It requires that as the sample size increases, the joint distribution of key variables (current treatments and outcomes, recent outcomes, covariates, and exposure vectors) converges to a well-defined limiting distribution. This condition guarantees that the experimental environment exhibits required stability in large-sample regimes.

\noindent
\textbf{Notation convention.} We fix a generic feasible treatment assignment $\Mobservedtreatment{\TH}$, and omit explicit references when the context is clear. We also employ the following notations for any~$t$:
\begin{align}
    \label{eq:notations}
    \left(
    \treatment{\UN}{t},
    \outcome{\UN}{t-1},
    \Vcovar{\UN}{t},
    \Vexposure{\UN}{t},
    \outcome{\UN}{t}
    \right)
    \sim
    \empdist{t;\Mobservedtreatment{t}}{\UN}
    \quad
    \text{and}
    \quad
    \left(
    \treatment{}{t},
    \outcome{}{t-1},
    \Vcovar{}{t},
    \Vexposure{}{t},
    \outcome{}{t}
    \right)
    \sim
    \limdist{t;\Mobservedtreatment{t}}{},
\end{align}
without explicitly restating the detailed conditions specified in Assumption \ref{asmp:empirical_distributions}.

Assumptions~\ref{asmp:unit-level_dynamics} and \ref{asmp:empirical_distributions} together with the definition of joint empirical distribution immediately imply that in finite samples where $\UN < \infty$, $\outcome{\UN}{t}$ is equal to
$\outcomefunction{}
\big(
\treatment{\UN}{t},
\outcome{\UN}{t-1},
\Vcovar{\UN}{t};
\Vexposure{\UN}{t}
\big)$.
This equality is straightforward since each
$\big(
\treatment{\UN}{t},
\outcome{\UN}{t-1},
\Vcovar{\UN}{t},
\Vexposure{\UN}{t},
\outcome{\UN}{t}
\big)$
take values with equal probability over the corresponding values of all units. Indeed, each
$\big(
\treatment{i}{t},
\outcome{i}{t-1},
\Vcovar{i}{t},
\Vexposure{i}{t},
\outcome{i}{t}
\big)$
will be observed under $\empdist{t}{\UN}$ with a probability of $\frac{1}{\UN}$. However, to analyze the limiting behavior as $\UN \rightarrow \infty$, we require additional theoretical considerations, as established in the following theorem.

\begin{theorem}[ESE-I]
    \label{thm:BSE-I}
    Under Assumptions~\ref{asmp:exposure_vecs}-\ref{asmp:empirical_distributions}, if the function~$\outcomefunction{}$ is continuous, we have:
    \begin{align}
        \label{eq:BSE-I}
        \outcome{}{t}
        \eqas
        \outcomefunction{}
        \left(
            \treatment{}{t},
            \outcome{}{t-1},
            \Vcovar{}{t};
            \Vexposure{}{t}
        \right)
    \end{align}
    where the recursion initiates from pretreatment outcome $\outcome{}{0}$ and $\eqas$ denotes almost sure equality.
\end{theorem}

The result of Theorem~\ref{thm:BSE-I} establishes the foundation for our subsequent analysis. The main objective of Theorem~\ref{thm:BSE-I} is to rigorously characterize the assumptions required to ensure we can transition from unit-level evolution specified by Eq.~\eqref{eq:outcome_specification} to the aggregate-level evolution outlined in Eq.~\eqref{eq:BSE-I}. However, directly approaching Eq.~\eqref{eq:BSE-I} presents significant challenges as the exposure vectors $\Vexposure{}{t}$, which capture the interference structure, are essentially black boxes. Below, we provide a concise proof for Theorem
\ref{thm:BSE-I}, and in the subsequent section, we conduct a detailed analysis of the elements of Eq.~\eqref{eq:BSE-I}.

\noindent
\textbf{Proof of Theorem~\ref{thm:BSE-I}.}
Let $\fgraph_{\outcomefunction{}}$ be the graph of the function $\outcomefunction{}$:
\begin{align*}
    \fgraph_{\outcomefunction{}}
    :=
    \Big\{
        (x,y):
        x \in \big(\treatmentspace \times \outcomespace \times \covarspace \times \exposurespace \big),\;
        y \in \outcomespace,\;
        y = \outcomefunction{}(x)
    \Big\}.
\end{align*}
Consider Assumptions~\ref{asmp:unit-level_dynamics} and \ref{asmp:empirical_distributions}, as well as the definition of empirical distributions, we know that
\begin{align*}
    \empdist{t;\Mobservedtreatment{t}}{\UN}
    \left(
        \fgraph_{\outcomefunction{}}
    \right)
    = 1.
\end{align*}
Since $\outcomefunction{}$ is a continuous function, the graph $\fgraph_{\outcomefunction{}}$ defines a closed set, and we can apply the Portmanteau theorem (e.g., Theorem 2.1 of \cite{billingsley2013convergence}) to get
\begin{align*}
    1
    =
    \limsup_{\UN \rightarrow \infty}\;
    \empdist{t;\Mobservedtreatment{t}}{\UN}
    \left(
        \fgraph_{\outcomefunction{}}
    \right)
    \leq
    \limdist{t;\Mobservedtreatment{t}}{}
    \left(
        \fgraph_{\outcomefunction{}}
    \right),
\end{align*}
which concludes the proof. \ep

\section{Interference Structure and Exposure Vectors}
\label{sec:interference}
In this section, we examine the exposure vector concept as specified by Assumption~\ref{asmp:exposure_vecs}. Basically, this assumption asserts that there exists a fixed-dimension vector that captures all relevant information from other units necessary to determine the potential outcome of each unit $i$. For example, the exposure vector $\Vexposure{i}{t}$ may take the form of neighborhood averages or network centrality measures, any of which provides a finite-length summary sufficient to model interference effects \citep{toulis2013estimation,basse2019randomization,forastiere2021identification,forastiere2022estimating,leung2022causal}.

However, in realistic settings, we typically face a partially known interference structure. Our goal is to revise the ESE equation (Eq.~\eqref{eq:BSE-I}) with this perspective.
We first explain the exposure mapping approach and decompose its concept into two distinct components: \emph{network structure} (how units are connected) and \emph{exposure mechanisms} (how those connected units influence each other).

\subsection{Exposure Mapping}
\label{sec:expo_mapping}
We proceed by adapting the exposure mapping definition from \cite{aronow2017estimating} to our setting by incorporating the temporal dimension.
\begin{definition}[Exposure mapping]
    \label{def:expo_mapping}
    Fix the treatment allocation $\Mobservedtreatment{\TH}$. The exposure mapping of round $t$ is a function $\expfunction{}{t}$ such that for all units $i$, we have
    \begin{align}
        \label{eq:exposure_mapping}
        \expfunction{}{t}: \treatmentspace^{\UN} \times \outcomespace^{\UN} \times \covarspace
        \mapsto
        \exposurespace,
        \quad
        \Vexposure{i}{t}(\Mobservedtreatment{t})
        :=
        \expfunction{}{t}
        \big(
            \Vobservedtreatment{}{t},
            \Voutcome{}{t-1}(\Mobservedtreatment{t-1}),
            \Vcovar{i}{t}
        \big).
    \end{align}
\end{definition}
Note that Definition~\ref{def:expo_mapping} specifies the exposure vectors while explicitly differentiates between immediate spillover effects (arising immediately from treatment assignment $\Vobservedtreatment{}{t}$) and peer effects that propagate through intermediate changes in units' outcomes $\Voutcome{}{t-1}(\Mobservedtreatment{t-1})$.

The primary purpose of the exposure mapping framework is to simplify subsequent analysis by reducing model dimensionality (ideally achieving $\exposuredim \ll \UN$). These mappings fundamentally encompass two distinct yet closely interrelated components of interference patterns:
\begin{itemize}
    \item \textbf{Network structure}: This component specifies the connectivity structure between units through either binary relationships (e.g., adjacency matrices indicating whether units are connected) or weighted relationships (e.g., weighted directed graphs quantifying the magnitude of unit interactions). In the framework of \citet{aronow2017estimating}, this structural information can be incorporated into Eq.~\eqref{eq:exposure_mapping} through the covariate vector $\Vcovar{i}{t}$. For instance, $\Vcovar{i}{t}$ may include the $i$-th row of a network adjacency matrix, thereby embedding unit $i$'s connectivity pattern.
    \item \textbf{Exposure mechanism}: This component describes the influence rules that govern how connected units affect one another's outcomes, essentially encoding the functional form of effects. Within the framework of \citet{aronow2017estimating}, these mechanisms are represented by the mathematical structure of the functions $\expfunction{}{t}$ in Eq.~\eqref{eq:exposure_mapping}. For example, \citet{cai2015social} employs a model where spillover effects are captured through the fraction of treated neighbors, implementing an averaging mechanism with respect to peers' treatment assignments.
\end{itemize}

To effectively leverage this framework, properly specified exposure mappings are essential \citep{aronow2017estimating}. Recent works study misspecified exposure mappings by restricting the scope of potential misspecification \citep{leung2022causal} and separating the role of exposure mappings in defining causal effects of interest from assumptions about interference structure \citep{savje2024causal}. Nevertheless, misspecification risk remains a substantial methodological challenge both at the exposure mechanism level (as discussed in \citet{auerbach2024discussion}) and through unobserved network connections \citep{egami2021spillover,weinstein2023causal}.%
\footnote{For insightful discussions on this topic, see the exchanges surrounding \citet{savje2024causal} in \citet{auerbach2024discussion,leung2024discussion,10.1093/biomet/asad071}.}

While domain knowledge can often guide the specification of exposure mechanisms, identifying network structures presents significant challenges. Many settings in public health, marketplace experimentation, and social science involve large populations of interacting units. Tracking network connections in such contexts would require close observation of each experimental unit throughout the entire experiment. In what follows, we demonstrate how causal message-passing provides a robust framework that bypasses the need for explicit network structure knowledge; thereby, we gain the required insight for handling the exposure vector $\Vexposure{i}{t}$ in our setting.

\subsection{Causal Message Passing (CMP)}
\label{sec:CMP}
We begin by presenting a simplified version of the CMP outcome evolution \citep{shirani2025can}. Let $g$ and $h$ be two real-valued functions defined on $\treatmentspace \times \outcomespace \times \covarspace$. Consider $\IMatl{ij}$ and $\IMatTl{ij}{t}$ as unknown weights that quantify the \emph{fixed} and \emph{time-varying} influence of unit $j$ on unit $i$, respectively. CMP considers the following family of outcome evolutions:
\begin{align}
    \label{eq:outcome_specification_CMP}
    \outcome{i}{t}(\Mobservedtreatment{t})
    :=
    h
    \big(
    \observedtreatment{i}{t},
    \outcome{i}{t-1}(\Mobservedtreatment{t-1}),
    \Vcovar{i}{t}
    \big)
    +
    \sum_{j=1}^\UN
    (\IMatl{ij}+\IMatTl{ij}{t})
    \outcomeg{}
    \big(
    \observedtreatment{j}{t},
    \outcome{j}{t-1}(\Mobservedtreatment{t-1}),
    \Vcovar{j}{t}
    \big).
\end{align}
Considering the outcome evolution specified in Eq.~\eqref{eq:outcome_specification}, CMP posits an additive structure. Specifically, Eq.~\eqref{eq:outcome_specification_CMP} assumes that each unit's potential outcome evolves as the sum of two components: a unit-specific term (characterized by the $h$-part) and a weighted aggregation of all experimental units' status (characterized by $\IMatl{ij}$, $\IMatTl{ij}{t}$, and the $\outcomeg{}$-part). From the exposure mapping perspective, CMP employs a one-dimensional exposure vector defined as
\begin{align}
    \label{eq:CMP_as_ExpoMapping}
    \exposure{i}{t}(\Mobservedtreatment{t})
    :=
    \sum_{j=1}^\UN
    (\IMatl{ij}+\IMatTl{ij}{t})
    \outcomeg{}
    \big(
    \observedtreatment{j}{t},
    \outcome{j}{t-1}(\Mobservedtreatment{t-1}),
    \Vcovar{j}{t}
    \big).
\end{align}
In view of Eq.~\eqref{eq:exposure_mapping}, CMP indeed separates the two components of exposure mapping under an additive assumption. Specifically, in Eq.~\eqref{eq:CMP_as_ExpoMapping}, the weights $\IMatl{ij}+\IMatTl{ij}{t}$ capture the network structure, while the function $g$ represents the underlying exposure mechanism.

Considering Eq.~\eqref{eq:CMP_as_ExpoMapping}, CMP explicitly models how treatments trigger links. Note that the term
$(\IMatl{ij}+\IMatTl{ij}{t})
\outcomeg{}
\big(
\observedtreatment{j}{t},
\outcome{j}{t-1}(\Mobservedtreatment{t-1}),
\Vcovar{j}{t}
\big)$
describes the contribution of unit $j$ to the outcome of unit $i$ in round $t$. When unit $j$ is treated in round $t$, the functionality of the link is shifted by replacing the baseline contribution 
$(\IMatl{ij}+\IMatTl{ij}{t})
\outcomeg{}
\big(
0,
\outcome{j}{t-1}(\Mobservedtreatment{t-1}),
\Vcovar{j}{t}
\big)$
with the treatment-induced contribution
$(\IMatl{ij}+\IMatTl{ij}{t})
\outcomeg{}
\big(
1,
\outcome{j}{t-1}(\Mobservedtreatment{t-1}),
\Vcovar{j}{t}
\big)$.
If this shift is non-zero, indicating that unit $j$ impacts unit $i$, the resulting effect is incorporated into the outcome of unit~$i$ in the subsequent round.

CMP specializes to dense networks by modeling both first- and second-order interactions through independent Gaussian weights: $\IMatl{ij} \sim \Nc(\ime{ij}{}/\UN,\sigma^2/\UN)$ and $\IMatTl{ij}{t} \sim \Nc(\ime{ij}{t}/\UN,\sigma^2_t/\UN)$. CMP then characterizes the limiting distribution of the interference effects. Specifically, let $\exposure{}{t}(\Mobservedtreatment{t})$ follow the same distribution as the empirical distribution of $\exposure{1}{t}(\Mobservedtreatment{t}), \ldots, \exposure{\UN}{t}(\Mobservedtreatment{t})$ as $\UN \rightarrow \infty$ in Eq.~\eqref{eq:CMP_as_ExpoMapping}. Under certain regularity conditions, CMP establishes the following dynamics for the exposure effects:
\begin{align}
    \label{eq:expo_CMP}
    \exposure{}{t}(\Mobservedtreatment{t})
    \eqas
    \Tilde{g}_t(\treatment{}{t}, \outcome{}{t-1}, \Vcovar{}{t}, \imegt{}{t}),
\end{align}
where $\Tilde{g}_t$ is a function that depends on $\outcomeg{}$ (see Eq.~\eqref{eq:outcome_specification_CMP}) and the distributions of the first-order weights $\ime{ij}{}+\ime{ij}{t}$. Furthermore, $\imegt{}{t}$ is a Gaussian random variable that reflects the impact of the second-order Gaussian weights and $\treatment{}{t}, \outcome{}{t-1}, \Vcovar{}{t}$ are as defined in Eq.~\eqref{eq:notations}.

Although this result allows CMP to bypass the observation of network structure by taking $\ime{ij}{}$, $\ime{ij}{t}$, $\sigma$, and $\sigma_t$ as unknowns, it still requires a rough characterization of the exposure mechanism. Specifically, \cite{shirani2024causal} build their estimation by considering simple exposure mechanisms where the average of previous round outcomes, the average of current treatments, and their interaction term serve as proxies for interference patterns across units. \cite{shirani2025can} extend this result and allow consideration of a set of candidate exposure mechanisms. They then present a counterfactual cross-validation method to enable automatic selection of exposure mechanisms.

\subsection{Partially Known Interference Structure}
\label{sec:partially_known_networks}
Our goal is to revise the ESE mappings in Eq.~\eqref{eq:BSE-I} by modeling the known and unknown components of the interference structure while ensuring the tractability of subsequent analysis. For this purpose, we consider the elements of the exposure vector for unit $i$ in round $t$, as specified by Assumption~\ref{asmp:exposure_vecs}:
\begin{align}
    \label{eq:expo_vec_elements}
    \Vexposure{i}{t}(\Mobservedtreatment{t})
    :=
    \big(
        \exposure{i (1)}{t}(\Mobservedtreatment{t}),
        \exposure{i (2)}{t}(\Mobservedtreatment{t}),
        \ldots,
        \exposure{i (\exposuredim)}{t}(\Mobservedtreatment{t})
    \big)^\top.
\end{align}
In view of Definition~\ref{def:expo_mapping}, we consider a sequence of ``pseudo-functions'' $\expfunction{1}{t}, \ldots, \expfunction{\exposuredim}{t}$ such that
\begin{align}
    \label{eq:exposure_mapping_elements}
    \expfunction{l}{t}: \treatmentspace^{\UN} \times \outcomespace^{\UN} \times \covarspace
    \mapsto
    \R,
    \quad
    \exposure{i (l)}{t}(\Mobservedtreatment{t})
    :=
    \expfunction{l}{t}
    \big(
        \Vobservedtreatment{}{t},
        \Voutcome{}{t-1}(\Mobservedtreatment{t-1}),
        \Vcovar{i}{t}
    \big),
    \quad
    l = 1, \ldots, \exposuredim.
\end{align}
In Eq.~\eqref{eq:exposure_mapping_elements}, each $\expfunction{l}{t}$ defines an operator from current treatments and previous outcomes to a scalar exposure measure, incorporating unit covariates. We use the term pseudo-function to emphasize that each $\expfunction{l}{t}$ must remain well-defined when operating on infinite-dimensional vectors as $\UN \rightarrow \infty$.

Now, we can reflect the available information on interference structure through a careful design of pseudo-functions $\expfunction{l}{t}$. For example, suppose all we know is that the interaction network consists of $K$ distinct clusters, each representing a cohesive subgroup with stronger internal connections than external ones. Then, we let $\exposuredim = K$ and define $\expfunction{l}{t}$ to capture how units belonging to cluster $l$ impact unit $i$, with heterogeneity arising from unit-specific covariates $\Vcovar{i}{t}$. In the next step, inspired by CMP, we incorporate the unknown components in each direction of the exposure vector through the distribution of relevant treatments. In the clustered setting, we can assume that $\expfunction{l}{t}$ reflects the influence of cluster $l$ on unit $i$ through the distribution of treatments of units belonging to cluster~$l$.

Considering the random vector $\Vexposure{}{t}(\Mobservedtreatment{t})$ that follows the limiting empirical distribution of exposure vectors (as per Assumption
\ref{asmp:empirical_distributions}), we formalize the necessary conditions in the following assumption.
\begin{assumption}[Stable decomposition of interference]
    \label{asmp:well-defined_expo_mapping}
    For each round $t$ and any treatment allocation $\Mobservedtreatment{t}$, there exists a function $\expoSEfunction{t}$ and random objects $\RandomObj{t}(\Mobservedtreatment{t})$ and $\LatentRObj{t}$ such that
    $$\Vexposure{}{t}(\Mobservedtreatment{t}) \eqas \expoSEfunction{t}(\RandomObj{t}(\Mobservedtreatment{t}), \LatentRObj{t}),$$
    where the functions~$\expoSEfunction{t}$ and the random object $\LatentRObj{t}$ remain invariant across treatment allocations~$\Mobservedtreatment{t}$.
\end{assumption}
Assumption~\ref{asmp:well-defined_expo_mapping} is critical to our setting by decomposing the aggregate interference effects into three components: treatment-dependent factors $\RandomObj{t}(\Mobservedtreatment{t})$, latent interference components $\LatentRObj{t}$, and the functional relationship $\expoSEfunction{t}$ (which depends on pseudo-functions $\expfunction{l}{t}$). Specifically, the invariance condition (that both $\expoSEfunction{t}$ and $\LatentRObj{t}$ remain unchanged across treatment assignments) ensures that all variations in interference effects can be explained by random objects $\RandomObj{t}(\Mobservedtreatment{t})$.

\subsubsection{Illustrative examples}
\label{sec:examples}
We present several examples to clarify the conditions of Assumption~\ref{asmp:well-defined_expo_mapping} and explicitly characterize the treatment-endogenous object $\RandomObj{t}(\Mobservedtreatment{t})$ in each case.

\subsubsection*{Causal message passing.}
Considering Eq.~\eqref{eq:expo_CMP}, CMP provides a concrete example where Assumption~\ref{asmp:well-defined_expo_mapping} holds under dense interference structures, with $\RandomObj{t}(\Mobservedtreatment{t}) = \{\treatment{}{t}, \outcome{}{t-1}\}$, reflecting the distribution of current treatments and past outcomes. Notably, the additive model specification and second-order i.i.d. random weights enable a clean characterization of the exogenous component as $\LatentRObj{t} = \{ \Vcovar{}{t}, \imegt{}{t} \}$. While the covariate component $\Vcovar{}{t}$ may be partially observable, $\imegt{}{t}$ can be interpreted as unit-level noise in large-sample regimes~\citep{shirani2025can}.

\subsubsection*{Clustered network structure.}
Reconsider the clustered setting with $K$ clusters, and let $\treatment{(l)}{t}$ denote a random variable distributed as the treatments assigned to units in cluster $l$, for $l = 1, \ldots, K$. In this case,
$\RandomObj{t}(\Mobservedtreatment{t}) = \big\{ \treatment{(1)}{t}, \ldots, \treatment{(K)}{t} \big\}.$
Intuitively, the treatment-endogenous object reduces to the collection of cluster-level treatment distributions. Combined with the unknown operators $\expfunction{l}{t}$, whose structure is captured by $\expoSEfunction{t}$ in Assumption~\ref{asmp:well-defined_expo_mapping}, this highlights the fact that each unit may be affected differently by each cluster of units.

\subsubsection*{Mean-field interference mechanisms.}
Another setting where Assumption~\ref{asmp:well-defined_expo_mapping} holds is when interference is mediated by mean-field quantities. For instance, \cite{munro2021treatment} analyze a market where spillovers arise through equilibrium prices that match supply and demand. In general, we may consider a mean-field quantity $\MF{}{t}(\Mobservedtreatment{t})$ that mediates interference effects for each unit under treatment scenario $\Mobservedtreatment{t}$. In this case, the treatment-endogenous object is $\RandomObj{t}(\Mobservedtreatment{t}) := \{\MF{}{t}(\Mobservedtreatment{t})\}$.

Note that Assumption~\ref{asmp:well-defined_expo_mapping} can also be viewed through a mean-field perspective. Indeed, the treatment-driven object $\RandomObj{t}(\Mobservedtreatment{t})$ serves as the mean-field quantity that summarizes \emph{aggregate} interference effects. This is more general than standard mean-field models, which typically assume that \emph{unit-level} interference is mediated by mean-field quantities. A deeper investigation of this connection is left for future work, as its relevance is context-dependent and in some applications unit-level mean-fields might be sufficient \citep{wager2021experimenting,johari2022experimental,munro2021treatment}.

\subsubsection*{Social influencers.}
Consider an experiment on a social media platform involving a small set of influencers and a large population of regular users. Influencers are users with disproportionately high reach and engagement who can affect the behavior of many others. Because of their outsized impact, we track each influencer’s treatment status individually while summarizing the remaining users through an aggregate measure. Without loss of generality, index units so that the influencers correspond to $1$ through $\exposuredim-1$. We then define
$\RandomObj{t}(\Mobservedtreatment{t}) = \big\{ \treatment{1}{t}, \ldots, \treatment{\exposuredim-1}{t}, \treatment{}{t} \big\}$,
where the first $\exposuredim-1$ components capture individual influencer treatments, and the final component $\treatment{}{t}$ represents the treatment distribution across all remaining users.

\subsection{Revised Experimental State Evolution}
\label{sec:}
Finally, we revise the result of Theorem~\ref{thm:BSE-I}; considering the conditions of Assumption~\ref{asmp:well-defined_expo_mapping} on aggregate interference patterns, the following result is immediate.
\begin{theorem}[ESE-II]
    \label{thm:BSE-II}
    Under Assumptions~\ref{asmp:exposure_vecs}-\ref{asmp:well-defined_expo_mapping}, if the function $\outcomefunction{}$ is continuous, there exist random functions $\SEfunction{}{t}$, $t=1, \ldots, \TH$, such that
    \begin{align}
        \label{eq:BSE-II}
        \outcome{}{t}
        \eqas
        \SEfunction{}{t}
        \left(
            \treatment{}{t},
            \outcome{}{t-1};
            \RandomObj{t}(\Mobservedtreatment{t})
        \right),
    \end{align}
    where
    \begin{align}
        \label{eq:ESE_mappings}
        \SEfunction{}{t}
        \left(
            \treatment{}{t},
            \outcome{}{t-1};
            \RandomObj{t}(\Mobservedtreatment{t})
        \right)
        :=
        \outcomefunction{}
        \Big(
            \treatment{}{t},
            \outcome{}{t-1},
            \Vcovar{}{t};
            \expoSEfunction{t}
            \big(
                \RandomObj{t}(\Mobservedtreatment{t}),
                \LatentRObj{t}
            \big)
        \Big)
    \end{align}
    and the recursion initiates from pretreatment initial outcomes $\outcome{}{0}$.
\end{theorem}
By the definition in Eq.~\eqref{eq:ESE_mappings}, for each $t$, the ESE mapping $\SEfunction{}{t}$ depends on unit-level evolutions captured through $\outcomefunction{}$, interference functional structure reflected by $\expoSEfunction{t}$, and limiting distributions of covariates and exogenous interference components (captured by $\Vcovar{}{t}$ and $\LatentRObj{t}$, respectively).

Theorem~\ref{thm:BSE-II} characterizes three distinct components of the experimental system. First, the initial outcomes collected before any treatment intervention, whose limiting empirical distribution is captured by $\outcome{}{0}$. Second, the ESE mappings $\SEfunction{}{t}$ that encode the evolution mechanisms governing the response of the experimental system to the treatment assignment. Crucially, these first two components remain unchanged regardless of treatment allocation. Third, the variable inputs to the ESE mappings that change across different treatment scenarios.

This decomposition reveals a key insight: \emph{ESE mappings characterize parallel evolution patterns across treatment scenarios, where aggregate outcomes evolve under identical mechanisms but with scenario-specific inputs}. We explore this parallel structure in the next section.

\section{Estimation Strategy: Distributional Parallel Propagations}
\label{sec:DPT}
\begin{figure}
    \centering
    \includegraphics[width=\linewidth]{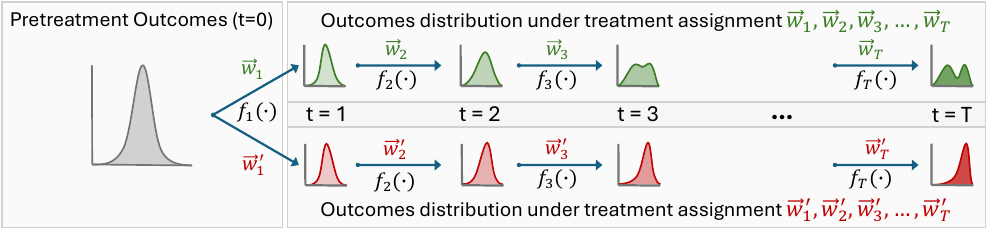}
    \caption{Functional parallelism: both scenarios evolve under same ESE mappings despite distinct distributions.}
    \label{fig:DPT}
\end{figure}

This section explains the parallel structures identified by ESE mappings across \emph{different treatment scenarios} and explains when they enable counterfactual estimation. To illustrate, consider two distinct treatment assignments $\Mobservedtreatment{\TH}$ and $\Mobservedtreatment{\TH}'$ and their corresponding panels of potential outcomes:
\begin{align}
    \label{eq:two_scenarios}
    \Moutcome{}(\Mobservedtreatment{\TH}) := \big[\Voutcome{}{0} \big| \Voutcome{}{1}(\Mobservedtreatment{1}) \big| \ldots \big| \Voutcome{}{\TH}(\Mobservedtreatment{\TH})\big]
    \quad\quad
    \Moutcome{}(\Mobservedtreatment{\TH}') := \big[\Voutcome{}{0} \big| \Voutcome{}{1}(\Mobservedtreatment{1}') \big| \ldots \big| \Voutcome{}{\TH}(\Mobservedtreatment{\TH}')\big].
\end{align}
From Theorem~\ref{thm:BSE-II} and Eq.~\eqref{eq:two_scenarios}, each outcome panel characterizes a distinct trajectory with specific structural properties. First, both trajectories originate from a \emph{common baseline}: the pretreatment observations $\Voutcome{}{0}$. Beginning in round $t=1$, these trajectories diverge into two distinct outcome vectors $\Voutcome{}{1}(\Mobservedtreatment{1})$ and $\Voutcome{}{1}(\Mobservedtreatment{1}')$. Crucially, while these outcomes differ, their distributions emerge from $\Voutcome{}{0}$ according to the same mathematical rule (the ESE mapping $\SEfunction{}{1}(\cdot)$) applied to different inputs $\Mobservedtreatment{1}$ and $\Mobservedtreatment{1}'$, respectively. This parallel evolutionary mechanism persists throughout the experiment timeline, generating two distinct but structurally parallel potential evolutions (Figure~\ref{fig:DPT}).

Now, suppose we observe outcomes under $\Mobservedtreatment{\TH}$ and aim to estimate the counterfactuals under $\Mobservedtreatment{\TH}'$. In the first stage, we use the observed outcomes $\Moutcome{}(\Mtreatment{\TH}{}=\Mobservedtreatment{\TH})$ to estimate the ESE mappings $\SEfunction{}{t}$. In the second stage, we exploit the parallel structure across treatment scenarios to construct counterfactual trajectories recursively. Starting from the common baseline of pretreatment outcomes $\Voutcome{}{0}$, we apply the estimated ESE mappings sequentially under the desired treatment assignment $\Mobservedtreatment{\TH}'$, which yields the aggregate counterfactual outcomes for the entire experimental population.

Similar evolution-based ideas appear in the difference-in-differences (DID) \citep{angrist2009mostly} and synthetic control methods (SCM) \citep{abadie2010synthetic}. DID relies on the \emph{parallel trends} assumption: the unobserved counterfactual trajectory of the treated group is assumed to evolve in parallel with the observed trajectory of the control group, starting from the pretreatment outcomes. SCM also uses pretreatment outcomes and constructs the counterfactual evolution of the treated group as a weighted combination of control group trajectories.

In general, parallel structures in the evolution of potential outcomes allow us to propagate forward from a common baseline and construct counterfactual trajectories through a sequential process. Under SUTVA, such parallelisms can be studied at the unit level, as in DID and SCM. Recent work extends this idea to networked settings by introducing modified parallel trends assumptions that account for interference \citep{xu2023difference,jetsupphasuk2025estimating}. However, our framework shifts the perspective to the aggregate level, where we model and characterize a conceptually similar form of functional parallelism at the distributional level.

Estimating the unknown ESE mappings $\SEfunction{}{t}$ is still challenging in practice. With only a single realization of the outcome panel, the information available for learning these mappings is limited. Focusing on treatment-induced variations provides new opportunities for estimation. In many applications, it is not necessary to recover the exact functional form of the ESE mappings, and suitable approximations may be sufficient. To formalize this idea, we consider two distinct treatment assignments: the baseline scenario with no treatment and $\Mobservedtreatment{\TH}$ representing an alternative assignment. Under Assumptions~\ref{asmp:empirical_distributions} and~\ref{asmp:well-defined_expo_mapping}, and using the notation established in Eq.~\eqref{eq:notations}, we adopt the following conventions for any~$t$:
\begin{equation}
    \label{eq:notations_two_scenarios}
    \begin{aligned}
        \left(
        0,
        \outcome{0}{t-1},
        \Vcovar{0}{t},
        \Vexposure{0}{t},
        \outcome{}{t}
        \right)
        \sim
        \limdist{t;\bf 0}{},
        \quad\quad\quad
        \Vexposure{0}{t} \eqas \expoSEfunction{t}(\RandomObj{t}^0, \LatentRObj{t}),
        \\
        \left(
        \treatment{}{t},
        \outcome{}{t-1},
        \Vcovar{}{t},
        \Vexposure{}{t},
        \outcome{}{t}
        \right)
        \sim
        \limdist{t;\Mobservedtreatment{t}}{},
        \quad\quad\quad
        \Vexposure{}{t} \eqas \expoSEfunction{t}(\RandomObj{t}, \LatentRObj{t}).
    \end{aligned}
\end{equation}
Here, the two sequences of random variables, denoted by $\outcome{0}{t}$ and $\outcome{}{t}$, represent the outcome distributions under their corresponding treatment scenarios.

\begin{theorem}
    \label{thm:counterfactuals_relation}
    Suppose that the conditions of Theorem~\ref{thm:BSE-II} hold and the ESE mappings $\SEfunction{}{t}$ are three times continuously differentiable for all $t$. Then,
    \begin{equation}
        \label{eq:counterfactuals_relation}
        \begin{aligned}
            \outcome{}{t}
            =
            \coeffW{t}
            \treatment{}{t}
            +
            \coeffY{t}
            \outcome{}{t-1}
            +
            \coeffI{t}
            \RandomObj{t}
            +
            \coeffWY{t}
            \treatment{}{t}
            \outcome{}{t-1}
            +
            \coeffWI{t}
            \treatment{}{t}
            \RandomObj{t}
            +
            \coeff{t}
            +
            R_t,
        \end{aligned}
    \end{equation}
    where $\coeffW{t}, \coeffY{t}, \coeffI{t}, \coeffWY{t}, \coeffWI{t}$, and $\coeff{t}$ are random coefficients depending on the baseline scenario (with no treatment) and independent of $\treatment{}{t}$. Additionally, the term $R_t$ collects the unretained second-order terms containing $(\outcome{}{t-1} - \outcome{0}{t-1})^2$, $(\RandomObj{t} - \RandomObj{t}^0)^2$, or $(\outcome{}{t-1} - \outcome{0}{t-1})(\RandomObj{t} - \RandomObj{t}^0)$, together with the third-order derivatives of $\SEfunction{}{t}$ evaluated at an intermediate point between the baseline and the perturbed inputs.
\end{theorem}
The relation in Eq.~\eqref{eq:counterfactuals_relation} captures how the treatment intervention, together with network effects, shifts the outcomes relative to the baseline scenario. When treatment effects are small in magnitude, the quantities $\outcome{}{t-1}$ and $\outcome{0}{t-1}$ (as well as $\RandomObj{t}$ and $\RandomObj{t}^0$) are expected to remain sufficiently close. In this case, the remainder term $R_t$ in Eq.~\eqref{eq:counterfactuals_relation} can be ignored. The smoothness assumption in Theorem~\ref{thm:counterfactuals_relation} is also standard in the literature, see for example \citet{li2022random}.

\noindent
\textbf{Proof of Theorem~\ref{thm:counterfactuals_relation}.}
By the result of Theorem~\ref{thm:BSE-II} and Assumption~\ref{asmp:well-defined_expo_mapping}, we know that
\begin{equation}
    \label{eq:proof_CR_ESE}
    \begin{aligned}
        \outcome{0}{t}
        \eqas
        \SEfunction{}{t}
        \left(
            0,
            \outcome{0}{t-1};
            \RandomObj{t}^0
        \right),
        \quad\quad
        \outcome{}{t}
        \eqas
        \SEfunction{}{t}
        \left(
            \treatment{}{t},
            \outcome{}{t-1};
            \RandomObj{t}
        \right),
    \end{aligned}
\end{equation}
where both recursions start from the same pre-treatment outcome $\outcome{}{0}$. Note that Eq.~\eqref{eq:proof_CR_ESE} involves two sequences of almost sure events. Since the union of a countable sequence of measure-zero sets is a measure-zero set, there exists $\Sc \subseteq \Omega$ with $\P(\Sc)=1$ such that for all $\omega \in \Sc$, the equalities in Eq.~\eqref{eq:proof_CR_ESE} hold for all $t$ (the dependence of random variables on $\omega$ is omitted for clarity of presentation). Applying Taylor's expansion pathwise for every $\omega \in \Sc$, we get the following:
\begin{equation}
    \label{eq:proof_CR_Taylor_1}
    \begin{aligned}
        \outcome{}{t}
        =
        &\;
        \outcome{0}{t}
        +
        \treatment{}{t}
        \partial_x
        \SEfunction{}{t}
        +
        (\outcome{}{t-1} - \outcome{0}{t-1})
        \partial_y
        \SEfunction{}{t}
        +
        (\RandomObj{t} - \RandomObj{t}^0)
        \partial_z
        \SEfunction{}{t}
        \\
        &\;
        + \frac{1}{2} (\treatment{}{t})^2 \partial_x^2 \SEfunction{}{t} %
        +
        \treatment{}{t}
        (\outcome{}{t-1} - \outcome{0}{t-1})
        \partial_{xy}
        \SEfunction{}{t}
        +
        \treatment{}{t}
        (\RandomObj{t} - \RandomObj{t}^0)
        \partial_{xz}
        \SEfunction{}{t}
        +
        R_t,
    \end{aligned}
\end{equation}
where $\partial_{x}\SEfunction{}{t}$, $\partial_{y}\SEfunction{}{t}$, and $\partial_{z}\SEfunction{}{t}$ denote the partial derivatives of $\SEfunction{}{t} \left(0, \outcome{0}{t-1}, \RandomObj{t}^0 \right)$ with respect to its first, second, and third arguments respectively; higher-order derivatives follow the same notation convention. In Eq.~\eqref{eq:proof_CR_Taylor_1}, the term $R_t$ collects the unretained second-order terms together with the third-order derivatives of $\SEfunction{}{t}$ as the remainder of our Taylor expansion. Considering $\treatment{}{t}$ is a 0-1 variable, we know that $(\treatment{}{t})^2=\treatment{}{t}$ and we can rewrite Eq.~\eqref{eq:proof_CR_Taylor_1} as given by Eq.~\eqref{eq:counterfactuals_relation} with 
\begin{equation*}
    \begin{aligned}
        \coeffW{t}
        &:=
        \partial_x
        \SEfunction{}{t}
        +
        \frac{1}{2} \partial_x^2 \SEfunction{}{t}
        -
        \outcome{0}{t-1}
        \partial_{xy}
        \SEfunction{}{t}
        -
        \RandomObj{t}^0
        \partial_{xz}
        \SEfunction{}{t},
        \\
        \coeffY{t}
        &:=
        \partial_y
        \SEfunction{}{t},
        \quad\quad
        \coeffI{t}
        :=
        \partial_z
        \SEfunction{}{t},
        \quad\quad
        \coeffWY{t}
        :=
        \partial_{xy}
        \SEfunction{}{t},
        \quad\quad
        \coeffWI{t}
        :=
        \partial_{xz}
        \SEfunction{}{t},
        \\
        \coeff{t}
        &:=
        \outcome{0}{t}
        - \outcome{0}{t-1}
        \partial_y
        \SEfunction{}{t}
        - \RandomObj{t}^0
        \partial_z
        \SEfunction{}{t}.
    \end{aligned}
\end{equation*}
This concludes the proof. \ep

The proof of Theorem~\ref{thm:counterfactuals_relation} reveals that randomized treatment assignment plays a fundamental role. Because treatments are assigned independently of all other factors, including the initial pre-treatment outcomes and so the baseline outcomes, we can clearly isolate treatment-induced variations (whether direct or network-mediated) in aggregate outcomes from the underlying model parameters. This separation enables simplifying the estimation of unknown ESE mappings to the estimation of the unknown coefficients in Eq.~\eqref{eq:counterfactuals_relation}. Once these coefficients are estimated, we can substitute $\treatment{}{t}$ in Eq.~\eqref{eq:counterfactuals_relation} with treatment assignments from alternative scenarios and, starting from the same pre-treatment outcomes $\outcome{}{0}$, recursively construct other counterfactual trajectories.

We emphasize that, for the coefficients in Eq.~\eqref{eq:counterfactuals_relation} to be identifiable, additional structural assumptions are needed. For example, in the CMP framework, where $\RandomObj{t} = \{\treatment{}{t}, \outcome{}{t-1}\}$, the right-hand side of Eq.~\eqref{eq:counterfactuals_relation} admits a simpler form. If we further assume that the coefficients are time-invariant, this formulation reduces to the model of \citet{shirani2024causal}, for which consistency results are available. For the other examples in \S~\ref{sec:examples}, Eq.~\eqref{eq:counterfactuals_relation} should be reformulated to incorporate the specific structural features of the interference mechanisms in each setting and to ensure that the resulting model remains identifiable. A full examination of this problem lies beyond the scope of this work and should be tailored to each setting based on the experimental context.

\section{Discussion and Conclusion}
\label{sec:conclusion}
This work adopts an evolution-based perspective on counterfactual estimation under interference. Instead of modeling outcomes directly as functions of treatments, we focus on evolutionary models that capture how outcomes transition from one observation to the next, characterizing the treatment-driven variations. Related approaches, such as Markovian interference models \citep{farias2022markovian} and switchback experiments \citep{hu2022switchback,bojinov2023design}, rely \emph{directly} on temporal outcomes to improve estimation and design. Our framework instead centers on the evolution rules: the mechanism that operates \emph{between} observations and captures how treatment-triggered interference drives aggregate changes in outcome dynamics.

Building on the causal message passing framework, which offers a concrete instantiation of evolution-based estimation \citep{shirani2024causal,shirani2025can}, we take a broader view. We investigate the minimal structural conditions under which evolution-based analysis is possible, albeit imposing stronger regularity assumptions on potential outcomes. We also highlight the notion of distributional parallel propagation as the functional parallelism across counterfactual scenarios. These results clarify how evolution-based approaches should be interpreted.

The core operating assumption requires a stable decomposition of aggregate interference effects into two parts: an endogenous component that varies across treatment scenarios and an exogenous component that remains invariant to treatment assignment. This decomposition enables isolating treatment-driven variations as outcomes evolve during the experiment. However, in certain settings, this invariance property is unrealistic. Exposure mechanisms may themselves shift with treatment. For example, in systems with threshold dynamics where peer influence activates only after a critical mass of the treatment adoption, the required decomposition does not hold.

Relying on treatment-driven variations over time also constrains the applicability of evolution-based frameworks. To illustrate this, we apply two versions of the CMP algorithm to the election-simulator data of \cite{shirani2025simulating}, which replicates the voter mobilization experiment of \cite{bond201261}. The simulator generates a synthetic social network and records voting intentions for 20{,}000 users beginning 40~days before election day. 

\begin{figure}[H]
    \centering
    \begin{subfigure}[b]{0.495\linewidth}
        \centering
        \subcaption*{\footnotesize Informational messages with ``weak'' treatment signal}
        \includegraphics[width=\linewidth]{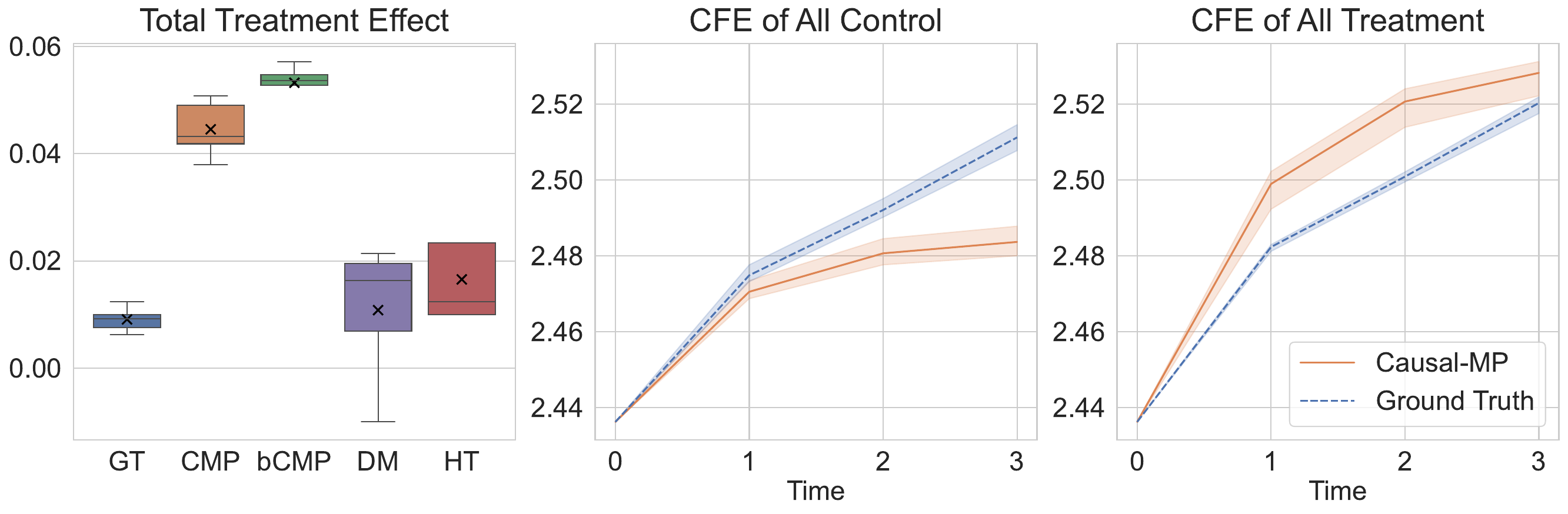}
    \end{subfigure}
    \hfill
    \begin{subfigure}[b]{0.495\linewidth}
        \centering
        \subcaption*{\footnotesize Social messages with ``strong'' treatment signal}
        \includegraphics[width=\linewidth]{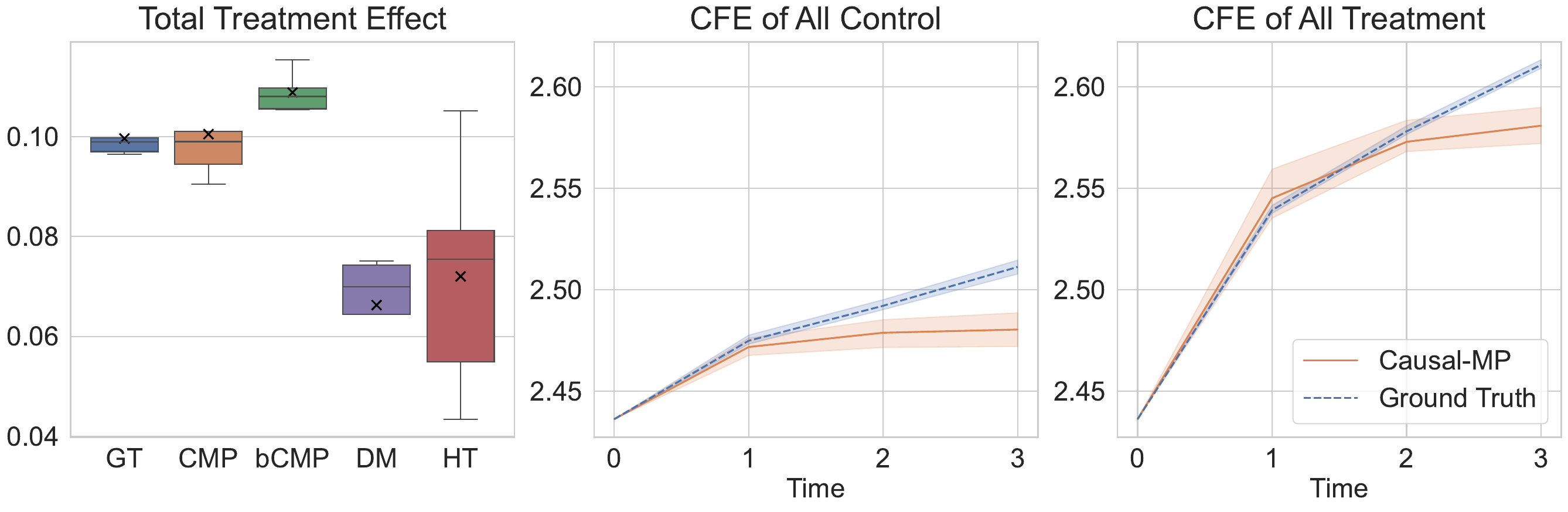}
    \end{subfigure}
    \caption{The performance of evolution-based estimators under strong time trends: CMP succeeds when the treatment signal is strong (social message) but struggles when it is weak (informational message).}
    \label{fig:combined-LLM}
\end{figure}

The experiment includes two treatment scenarios: an informational message that provides factual voting details and a social message that additionally displays peer behavior. Each user is independently assigned treatment with probability 0\% in the first 10 days and with probabilities 20\%, 40\%, and 80\% in the second, third, and fourth 10-day blocks, respectively. Because users are not online every day, the raw data contain missing outcomes. We therefore aggregate each 10-day block into a single round, producing a panel with four rounds. Since the simulator is fully controlled, both scenarios provide ground-truth outcomes, enabling evaluation of the estimation method.

Figure~\ref{fig:combined-LLM} reports the results, comparing ground-truth (GT) values with estimates from basic CMP (bCMP) \citep{shirani2024causal}, the full CMP method \citep{shirani2025can}, the difference-in-means estimator (DM), and the Horvitz–Thompson estimator (HT)%
\footnote{Difference-in-means (DM) and Horvitz-Thompson (HT) are expressed as:
\begin{align*}
\DIME :=
\frac{\sum_{i=1}^N\outcomeD{}{i}{\TH}\treatment{i}{\TH}}{\sum_{i=1}^N\treatment{i}{\TH}} - \frac{\sum_{i=1}^N\outcomeD{}{i}{\TH}(1-\treatment{i}{\TH})}{\sum_{i=1}^N(1-\treatment{i}{\TH})},
\quad\quad
\HTE :=  \frac{1}{N} \sum_{i=1}^N \left( \frac{\outcomeD{}{i}{\TH} \treatment{i}{\TH}}{\E[\treatment{i}{\TH}]} - \frac{\outcomeD{}{i}{\TH} (1 - \treatment{i}{\TH})}{\E[1 - \treatment{i}{\TH}]} \right).
\end{align*}}
\citep{savje2021average}. For each scenario, we also plot the counterfactual evolution of the sample-mean outcomes (CFE) under both all-control and all-treatment conditions, using the GT values and the CMP estimates.

In both scenarios, the middle panels show a clear upward trend in the all-control case, reflecting the natural increase in voting intention as election day approaches. In the informational-message scenario (left panels), where the treatment signal is weak, both CMP methods struggle to separate treatment-induced variation from this strong time trend and therefore overestimate the treatment effect. In contrast, in the social-message scenario (right panels), where the treatment signal is substantially stronger, both CMP methods succeed in distinguishing treatment-driven changes from the underlying time trend and produce improved treatment-effect estimates.

In conclusion, in settings where network interference is widespread, it may not be practical to attempt direct observation of the underlying network. Evolution-based approaches are particularly useful in such cases, but their validity depends on the interference structures remaining stable across treatment allocations. When this stability holds, methods such as causal message passing algorithms offer a principled and cost-effective strategy for estimating treatment effects.

\newpage

\bibliography{mypaper}
\bibliographystyle{apalike}

\end{document}